\documentclass[10pt, a4paper]{article}

\usepackage{lrec2026} 
\usepackage{booktabs} 
\usepackage{amsmath}
\usepackage{pifont}
\usepackage{subcaption}
\usepackage{multirow}
\usepackage{caption}
\usepackage{tikz} 
\usepackage{float}  

\title{Evaluation of Pose Estimation Systems for Sign Language Translation}

\name{Catherine O'Brien*\thanks{* These authors contributed equally.}, Gerard Sant*, Mathias Müller, Sarah Ebling} 

\address{Department of Computational Linguistics \\
         University of Zurich, Switzerland \\
         \{catherineelizabeth.obrien, gerard.santmuniesa\}@uzh.ch, \{mmueller, ebling\}@cl.uzh.ch\\}

\date{
\textsuperscript{*}These authors contributed equally.
}

\abstract{
Many sign language translation (SLT) systems operate on pose sequences instead of raw video to reduce input dimensionality, improve portability, and partially anonymize signers. The choice of pose estimator is often treated as an implementation detail, with systems defaulting to widely available tools such as MediaPipe Holistic or OpenPose. We present a systematic comparison of pose estimators for pose-based SLT, covering widely used baselines (MediaPipe Holistic, OpenPose) and newer whole-body/high-capacity models (MMPose WholeBody, OpenPifPaf, AlphaPose, SDPose, Sapiens, SMPLest-X). We quantify downstream impact by training a controlled SLT pipeline on RWTH-PHOENIX-Weather 2014 where only the pose representation varies, evaluating with BLEU and BLEURT.
To contextualize translation outcomes, we analyze temporal stability, missing hand keypoints, and robustness to occlusion using higher-resolution videos from the Signsuisse dataset. SDPose and Sapiens achieve the best translation performance (BLEU \textasciitilde 11.5), outperforming the common MediaPipe baseline (BLEU \textasciitilde 10). In occlusion cases, Sapiens is correct in all tested instances (15/15), while OpenPifPaf fails in nearly all (1/15) and also yields the weakest translation scores. Estimators that frequently leave out hand keypoints are associated with lower BLEU/BLEURT. We release code that can be used not only to reproduce our experiments, but also considerably lowers the barrier for other researchers to use alternative pose estimators.
 \\ \newline  }

\begin{document}

\maketitleabstract

\section{Introduction}\label{sec:introduction}


Sign language processing (SLP) is gaining ground within Natural Language Processing (NLP), yet it remains substantially underrepresented \citep{bragg_etal_2019_signlanguagerecognition, yin-etal-2021-including, muller-etal-2022-findings}. In spoken-language NLP, many core modeling decisions and preprocessing choices have been systematically studied and benchmarked. In contrast, for sign languages, even fundamental design decisions remain underexplored. For instance, while tokenizer selection in spoken-language translation is informed by extensive prior work, it remains unclear how sign language data should best be preprocessed for sign language translation (SLT).

 
SLT systems can accept input from a variety of modalities, including raw video \citep{DBLP:conf/iccv/ZhouCC0LE0Z23, DBLP:conf/nips/Ye0J0X24,modality_matters}, glosses—semantic transcriptions of meaning— \citep{Camgoz_2018_CVPR, yin-read-2020-better, Zhou_2021_CVPR}, form notations like SignWriting \citep{sutton1990lessons, jiang-etal-2023-machine} or HamNoSys \citep{prillwitz1990hamnosys}, which encode articulatory form, or poses \citep{zhang2024scaling, 10.1371/journal.pone.0316298, DBLP:journals/jcst/GanYJXL25, 10943439}. Poses are approximations of a signer’s skeleton and movements as they sign.

Poses are a common choice for SLT because they are more viable than glosses or form notations \citep{muller-etal-2023-considerations}, and offer a lightweight interpretable alternative to raw video. They also provide some degree of anonymization when compared to signer videos \citep{battisti2024personidentificationfromposeestimates, moryossef-etal-2025-pose}. Poses may also be more signer-independent for low-resource sign languages \citep{holmes-etal-2022-improving}. Despite these advantages, video-based models still outperform pose-based systems \citep{Moryossef_2021_CVPR, Tarres_2023_CVPR, modality_matters}, suggesting that current pose representations may lose critical linguistic information. Improving pose estimation quality therefore represents a promising pathway toward narrowing this performance gap. 

Although numerous pose estimation systems can be applied to sign language data, their impact on downstream translation performance remains largely unexplored. In practice, most translation pipelines rely on MediaPipe \citep{lugaresi-etal-2019-mediapipe} or OpenPose \citep{cao-etal-2019-openpose}, primarily due to accessibility and ease of integration rather than demonstrated suitability for sign language. 

To our knowledge, this work presents the first systematic evaluation of pose estimators for SLT. Beyond translation performance, we examine their behavior under pose estimation challenges intrinsic to signing, including occlusion, temporal instability, and missing hand detections. Our analysis provides practical guidance for selecting pose representations and highlights factors that influence downstream translation quality.

The main contributions of this paper are: 
\begin{itemize}
    \item A controlled comparison of eight prominent pose estimators on the RWTH-PHOENIX-Weather 2014 dataset,
    \item An analysis of pose estimation failure modes relevant to sign language data, including occlusion, temporal instability, and missing hands keypoints,
    \item An open-source framework extending the \texttt{pose-format} library \citep{moryossef2023poseformatlibraryviewingaugmenting} with support for all evaluated pose estimators, simplifying integration and lowering barriers to their use in SLT pipelines. 
\end{itemize}

Code and scripts to reproduce our experiments are available at \url{https://github.com/ZurichNLP/multimodalhugs-pipelines}.

\section{Background}




\subsection{Pose Estimation} \label{subsec:pose_estimation}

Pose estimation systems extract human skeletal keypoints from video, representing body, hand, and facial articulators as spatio-temporal trajectories. Modern pipelines typically rely on deep learning–based detectors such as OpenPose \citep{cao-etal-2019-openpose}, MediaPipe Holistic \citep{lugaresi-etal-2019-mediapipe, grishchenko2020mediapipe}, or SMPL-based models \citep{10.1145/2816795.2818013, DBLP:conf/nips/CaiYZWSYPMZZLYL23, yin2025smplest} to recover 2D or 3D joint configurations from RGB input. These representations provide a compact and interpretable encoding of human motion and are widely used in sign language processing \citep{Wei_2016_CVPR, LI_2020_WACV, Saunders_2022_CVPR, joshi-etal-2025-posestitch} and generation pipelines \citep{DBLP:conf/eccv/SaundersCB20, DBLP:journals/nn/XiaoQY20, moryossef-etal-2023-open, Arkushin_2023_CVPR}.

In sign language contexts, accurate capture of fine-grained articulations—including handshape, orientation, movement, and facial expression—is essential, as linguistic meaning is conveyed through coordinated manual and non-manual signals \citep{liddell1989american, JohnsonLiddell2010PhoneticSigns, PfauQuer2010Nonmanuals, brentari2011sign, Sandler2012PhonologicalOrganization, 10.1371/journal.pone.0086268, yin-etal-2021-including}. Errors in keypoint localization or tracking may therefore lead to the loss of critical articulatory information \citep{Moryossef_2021_CVPR, moryossef2024optimizinghandregiondetection}, degrading the quality of representations used by downstream tasks.

Despite these challenges, sign language processing pipelines commonly rely on general-purpose pose estimators that are not optimized for sign language data, as noted in \citep{decoster2023extractionrobustsignembeddings, jiang-etal-2025-meaningful}. Recent advances in whole-body and 3D pose estimation offer improved hand and body modeling \citep{10.1007/978-3-030-58545-7_12, Zhu_2023_ICCV}. While pose-based representations are widely used in SLT, the implications of estimator choice for translation quality remain insufficiently understood.

\subsection{Sign language translation} \label{subsec:slt}

Sign language translation (SLT) seeks to generate spoken-language text from signed input \citep{muller-etal-2022-findings, DeCoster2024SignedMT}. SLT systems have employed a wide range of input representations, spanning continuous visual signals—such as raw video \citep{DBLP:conf/iccv/ZhouCC0LE0Z23, DBLP:conf/nips/Ye0J0X24} or learned visual features \citep{Tarres_2023_CVPR, gueuwou-etal-2025-shubert}, among others—as well as discrete representations including gloss annotations \citep{Camgoz_2018_CVPR, yin-read-2020-better, Zhou_2021_CVPR} and form notations \citep{jiang-etal-2023-machine}.

While raw video inputs often achieve the strongest translation performance among modalities \citep{modality_matters}, pose-based representations have been widely adopted in SLT pipelines \citep{zhang2024scaling, 10.1371/journal.pone.0316298, DBLP:journals/jcst/GanYJXL25, 10943439}, as they provide a lower-dimensional alternative that preserves articulatory structure. However, the effectiveness of pose-based SLT may be limited by the fidelity of the underlying pose representations, which varies across commonly used general-purpose estimators (Section~\ref{subsec:pose_estimation}).

Comparative studies of pose estimators for sign language tasks remain limited and report mixed findings. Several works report stronger performance of MediaPipe relative to OpenPose \citep{Moryossef_2021_CVPR, muller-etal-2022-findings, muller2023d4}, a trend further confirmed by \citet{decoster2023extractionrobustsignembeddings}, who also found MediaPipe to outperform MMPose \citep{jin2020whole}. In contrast, \citet{lazoquispe2022impactofposeestimationmodelsforlandmarkbasedslr} reported improved results using an MMPose-based pipeline. The impact of newer whole-body and high-capacity estimators on SLT performance has yet to be systematically examined.

\section{Pose Estimators} 

\begin{table*}
    \centering
    \footnotesize
    \resizebox{\textwidth}{!}{
    \begin{tabular}{lrrcrrc}
         \toprule
         \textbf{estimator} & \textbf{\# keypoints} & \textbf{scheme} & \textbf{2D/3D} & \textbf{speed (fps)} & \textbf{GPU only} & \textbf{confidence} \\
         \hline
    MediaPipe Holistic \citep{lugaresi-etal-2019-mediapipe} & $576$ & - & 3D & $0.89$ / $3.15$ & - & \ding{52} \\
    OpenPose \citep{cao-etal-2019-openpose} & $137$ & - & 2D & - / $4.40$ & \ding{52} & \ding{52} \\
    MMPose Wholebody \citep{jin2020whole} & $133$ & COCO Wholebody & 2D & $0.89$ / $3.81$ & - & \ding{52} \\
    OpenPifPaf \citep{kreiss2021openpifpaf} & $133$ & COCO Wholebody & 2D & $1.21$ / $4.42$ & - & \ding{52} \\
    SDPose \citep{liang2025sdposeexploitingdiffusionpriors} & $133$ & COCO Wholebody & 2D & $0.07$ / $0.84$ & - & \ding{52} \\
    Sapiens \citep{khirodkar2024sapiens} & $308$ & Sapiens-308 & 2D & $0.04$ / $3.29$ & - & \ding{52} \\
    AlphaPose \citep{alphapose} & $136$ & Halpe-FullBody & 2D & - / $22.89$ & \ding{52} & \ding{52} \\
    SMPLest-X \citep{yin2025smplest} & $137$ & SMPL-X & 2D & - / $8.36$ & \ding{52} & -\\
         \bottomrule
    \end{tabular}
    }
    \caption{General characteristics of pose estimators considered in this study. scheme=whether estimated keypoints correspond to a standard layout, speed=frames per second on CPU/GPU measured on a single V100, confidence=whether the estimator outputs confidence values for each keypoint}
    \label{tab:estimators_comparison}
\end{table*}


As shown in Table~\ref{tab:estimators_comparison}, we consider both pose estimators widely used in previous sign language processing research—such as MediaPipe \citep{lugaresi-etal-2019-mediapipe} and OpenPose \citep{cao-etal-2019-openpose}—as well as more recent systems that have not been extensively used but appear to be strong candidates. All evaluated methods are human pose estimators rather than models specialized for sign language. We selected both top-down and bottom-up approaches to pose estimation.
We exclude common pose estimators that do not estimate hands, such as YOLO \citep{maji2022yolo} or PoseFormer \citep{zheng20213d}.


\paragraph{Mediapipe Holistic \citep{lugaresi-etal-2019-mediapipe}}

This system employs a unified architecture to jointly estimate body pose, facial landmarks, and detailed hand keypoints by integrating region-specific subnetworks within a graph-based perception pipeline. This design produces dense $540$-landmark representations and enables coordinated tracking of upper-body articulators from RGB input, making it widely used in real-time gesture and sign language processing. Its emphasis on low-latency, on-device inference has further contributed to its widespread adoption in SLT pipelines \citep{zhang2024scaling, gueuwou-etal-2025-shubert}. In our experiments we reduce the full mesh of face keypoints to a smaller selection of contour points.

\paragraph{OpenPose \citep{cao-etal-2019-openpose}}

Building on Part Affinity Fields (PAFs) \citep{Cao_2017_CVPR}, which represents limb associations as 2D vector fields linking detected joints, this approach enables bottom-up multi-person pose estimation by jointly detecting keypoints and learning their pairwise associations. This formulation removes the dependency on person detection bounding boxes and improves robustness in crowded scenes. OpenPose further extended the pose estimation paradigm to full-body configurations including face and hands, which made it influential in multimodal and sign language research \citep{app9132683, electronics9081257, DBLP:conf/eccv/MoryossefTAEN20, electronics12122678, NUNEZMARCOS2023118993}.

We run a pre-built OpenPose docker image with \texttt{{-}-hands} and \texttt{{-}-face} enabled, resulting in a total of 137 keypoints per frame. Occasionally OpenPose incorrectly predicts several people, in which case we select the first one.

\paragraph{MMPose Wholebody \citep{jin2020whole}}


Extends conventional human pose estimation beyond standard $17$-keypoint body configurations to dense full-body representations, typically comprising 133 keypoints (including body, hands, face, and feet) as defined in the COCO-WholeBody benchmark \citep{jin2020whole}. Architecturally, it follows a top-down paradigm: person instances are first detected, after which high-resolution keypoint heatmaps are predicted for each region.


We do estimation for MMPose Wholebody via their Unified Inference API. This API offers several checkpoint options and a variety of outputs. "Wholebody" refers to the checkpoint \texttt{rtmpose-m\_simcc-coco-wholebody\_pt-aic-coco\_270e-256x192-cd5e845c\_20230123}, which outputs the 133 keypoints defined in the COCO-Wholebody dataset. 

\paragraph{OpenPifPaf \citep{kreiss2021openpifpaf}}

A bottom-up multi-person pose estimator builds on the Part Affinity Fields paradigm to jointly detect and associate keypoints. Its representation improves keypoint localization and grouping under partial occlusion and scale variation. OpenPifPaf has demonstrated strong robustness in crowded scenes \citep{Andriluka_2018_CVPR} while remaining computationally efficient. We estimate poses for OpenPifPaf using their Predictor API with the pretrained checkpoint \texttt{shufflenetv2k30-wholebody}. We estimate poses for OpenPifPaf using their Predictor API with the pretrained checkpoint \texttt{shufflenetv2k30-wholebody}. 

\paragraph{SDPose \citep{liang2025sdposeexploitingdiffusionpriors}}

Proposes a pose estimation framework that exploits pre-trained generative priors from Stable Diffusion \citep{Rombach_2022_CVPR} to enhance both standard accuracy and robustness under domain shift.
Evaluated against both in-domain and out-of-distribution benchmarks \citep{10.1007/978-3-030-58545-7_12, Ju_2023_CVPR} (e.g., human and stylized images), SDPose achieves competitive results with strong cross-domain generalization, highlighting the potential of diffusion-based priors in structured vision tasks such as pose estimation.

For SDPose, we use a pretrained model obtained from Hugging Face \texttt{teemosliang/SDPose-Wholebody}. This architecture first computes bounding boxes for each person using \texttt{YOLO-11x}. We have adapted their code such that we use only the person with the highest confidence, and discard any other persons detected. 

\paragraph{Sapiens \citep{khirodkar2024sapiens}}

Represents a shift toward large-scale pretraining for human-centric vision tasks. Rather than optimizing narrowly for 2D keypoint detection, it leverages foundation-model-style pretraining on large-scale human motion data and supports fine-tuning for pose, segmentation, and depth estimation. Its dense $308$-keypoint representation—including $243$ facial landmarks—and high-resolution inference enable precise capture of fine-grained articulations relevant to sign language (See Section~\ref{subsec:pose_estimation}). This paradigm aligns with trends in NLP, where large-scale pretraining improves transferability to downstream multimodal tasks.

We estimate Sapiens poses based on the implementation of \citet{Gorordo2024sapienspytorch}, which reproduces the official Sapiens pose model without requiring the full framework. We use the \texttt{sapiens\_1b\_goliath\_best\_goliath\_AP\_640} checkpoint from the official release; following the authors’ recommendations, the 1B model is preferred as smaller variants yield lower accuracy.

\paragraph{AlphaPose \citep{alphapose}}

A top-down multi-person pose estimator that performs human detection followed by pose regression within each bounding box. It improves localization accuracy through Symmetric Integral Keypoint Regression, enabling continuous joint prediction beyond discrete heatmap representations. AlphaPose propose the Halpe-FullBody scheme, which supports whole-body estimation (up to 136 keypoints), extending COCO-WholeBody format by including additional head, neck, and hip joints. Its multi-stage pipeline enables efficient inference while preserving localization accuracy.

We use the authors’ recommended \texttt{multi}- \texttt{\_domain\_fast50\_dcn\_combined\_256x192} checkpoint for Halpe-FullBody pose estimation.

\paragraph{SMPLest-X \citep{yin2025smplest}}

A minimalist one-stage model for expressive human pose and shape estimation based on the SMPL-X scheme \citep{Pavlakos_2019_CVPR} that jointly predicts body, hand, and face parameters using a transformer encoder–decoder architecture with task tokens, eliminating part-specific modules used in prior pipelines \citep{Rong_2021_ICCV, pymafx}. Designed to study scaling effects rather than architectural complexity, it is trained on large, diverse multi-dataset mixtures and shows strong cross-domain generalization. Despite its simplicity, the authors report state-of-the-art performance across multiple benchmarks \citep{Pavlakos_2019_CVPR, Patel_2021_CVPR, DBLP:conf/eccv/ZhangMZQKPBT22, Lin_2023_CVPR, DBLP:conf/cvpr/FanTTKKBH23} and highlight strong results for articulated hand pose estimation.

We use the \texttt{SMPLest-X-Huge} model recommended by the authors. Because this model does not provide per-keypoint confidence scores, we assume confidence $1$
to retain all predicted keypoints during training (Section~\ref{subsec:translation_method}).

\paragraph{Support for \texttt{pose-format}}

We provide instructions and code\footnote{\url{https://github.com/ZurichNLP/video-to-pose}} for running all of these lesser-used (in the context of SLT) estimators and also extend the \texttt{pose-format} library \citep{moryossef2023poseformatlibraryviewingaugmenting} to support these new pose types. This considerably lowers the barriers for other researchers to use alternative estimators. Compatibility with pose-format means the predicted poses can more easily be stored, loaded, manipulated and visualized.

\section{Methodology of Experiments}



This study compares multiple pose estimators in the context of sign language translation. We evaluate each estimator within an identical translation pipeline to measure its impact on translation quality (Section \ref{subsec:translation_method}). To contextualize these results, we also analyze estimator behavior with respect to temporal instability, occlusion, and missing hand detections—common challenges in pose estimation (Section \ref{sub:methodology_qualitative_analyses}).

\subsection{Translation} \label{subsec:translation_method}

We train translation models with the recent MultiModalHugs toolkit \citep{sant2025multimodalhugs}. MultiModalHugs is a framework built on top of Hugging Face designed for multimodal AI models, which makes it ideal for a dataset of .pose files. 

\textbf{Dataset} We use the RWTH-PHOENIX-Weather 2014 dataset \citep{forster-etal-2014-extensions, Camgoz_2018_CVPR}\footnote{Referred to as \textit{Phoenix} in the remainder of the text.}, which contains videos of German Sign language (DGS) aligned with German transcripts.

\textbf{Preprocessing} We process each video with each pose estimator to extract frame-level skeletal keypoints and store them in the binary .pose standard defined by the \texttt{pose-format} library.
This enables consistent loading of pose sequences of any type within MultiModalHugs. The resulting representations preserve keypoint coordinates and confidence scores (if available) for each frame, while accommodating estimator-specific keypoint layouts (Table~\ref{tab:estimators_comparison}).
Pose files are preprocessed as follows: leg keypoints are removed, and the remaining ones are spatially normalized per sequence to ensure consistent body and hand scale across signers. Missing values are zero-filled, and each frame is flattened into a feature vector. 

\begin{figure*}[!t]
    \centering
    \begin{subfigure}[t]{0.18\textwidth}
        \centering
        \includegraphics[width=\linewidth]{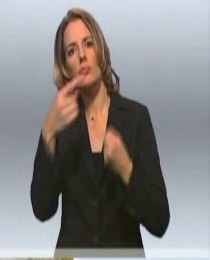}
        \caption{Original frame}
        \label{fig:orig}
    \end{subfigure}
    \hfill
    \begin{subfigure}[t]{0.18\textwidth}
        \centering
        \includegraphics[width=\linewidth]{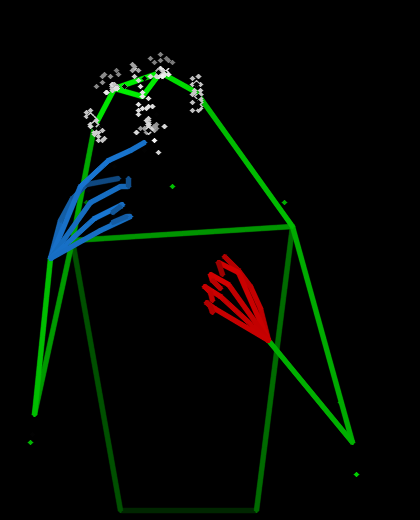}
        \caption{Sapiens}
        \label{fig:sapiens}
    \end{subfigure}
    \hfill
    \begin{subfigure}[t]{0.18\textwidth}
        \centering
        \includegraphics[width=\linewidth]{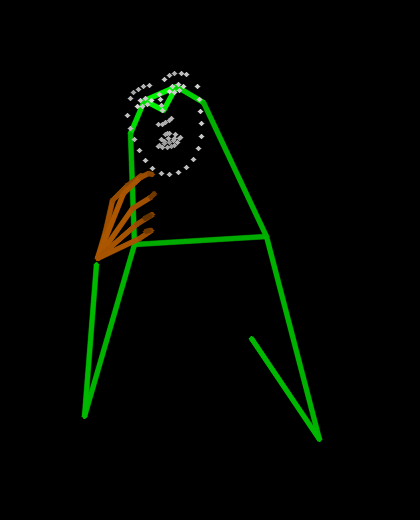}
        \caption{OpenPifPaf}
        \label{fig:openpifpaf_one}
    \end{subfigure}
    \hfill
    \begin{subfigure}[t]{0.18\textwidth}
        \centering
        \includegraphics[width=\linewidth]{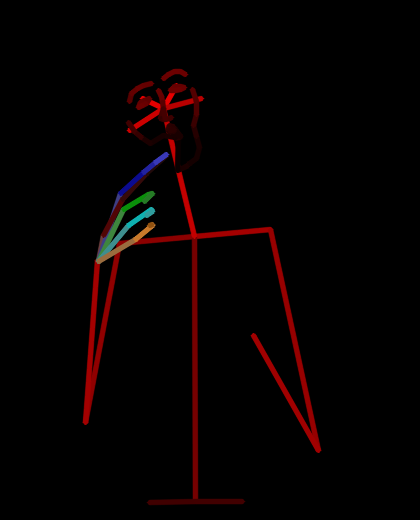}
        \caption{OpenPose}
        \label{fig:openpifpaf_two}
    \end{subfigure}
    \hfill
    \begin{subfigure}[t]{0.18\textwidth}
        \centering
        \includegraphics[width=\linewidth]{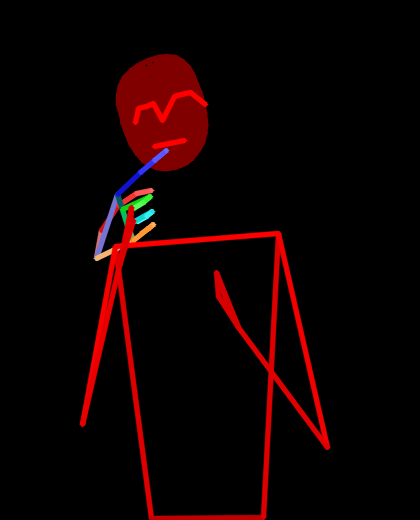}
        \caption{MediaPipe}
        \label{fig:mediapipe}
    \end{subfigure}
    
    
    \caption{Example illustrating missing hand keypoints across pose estimators. (a) Original RGB frame used to extract poses, shown for reference. (b) Sapiens predicts a complete set of hand keypoints (no confidence $=0$), although some finger articulation may still be inaccurate (e.g., the left index finger). (c–e) OpenPifPaf, OpenPose, and MediaPipe assign confidence $=0$ to large portions of the hand keypoints (complete for c–d and nearly complete for e). During preprocessing, these keypoints are masked and zero-filled, resulting in the loss of hand information provided to the translation model.}
    \label{fig:pose_comparison}
\end{figure*}

We employ a shared encoder–decoder architecture, following the pose-based pipeline described in \citet{modality_matters}. Pose sequences are represented as frame-level keypoint vectors and projected through a linear layer into the embedding space of a pretrained multilingual Transformer (\texttt{facebook/m2m100\_418M}). This adapter aligns modality-specific representations with the language model’s input format, ensuring that only the pose estimator varies across experiments.
The input dimensionality of each model is determined by the number of keypoints per estimator, multiplied by their spatial dimensionality (2D or 3D).

\textbf{Evaluation} To evaluate translation quality, we report BLEU \citep{papineni-etal-2002-bleu} using SacreBLEU \citep{post-2018-call}\footnote{Signature: \texttt{nrefs:1|case:mixed|eff:no|} \texttt{tok:13a|smooth:exp|version:2.6.0}} and BLEURT \citep{sellam2020bleurt,pu2021learning} using the \texttt{BLEURT-20} model.
Other standard evaluation metrics such as COMET \citep{rei-etal-2020-comet} are not applicable to our task, since they do not support sign languages.



\subsection{Further Analyses} \label{sub:methodology_qualitative_analyses}



We conduct the following quantitative or qualitative analyses: an investigation of temporal jitter, robustness to occlusion and missing hand keypoints.

\textbf{Datasets} We use both the Phoenix and the Signsuisse dataset \citep{muller-etal-2023-findings}. We add Signsuisse data because it includes higher-resolution videos than Phoenix and features deaf signers, whereas Phoenix has hearing interpreters. Besides, in order to accurately assess how estimators are handling occlusion, high-quality videos are required to reduce the influence of blur. We use the portion of the Signsuisse data that contains Swiss German Sign Language (\emph{Deutschschweizerische Gebärdensprache}, or DSGS) videos.

\paragraph{Temporal jittering (pose stability)}
Beyond visual inspection, we quantify the temporal stability of each pose estimator using derivative-based smoothness metrics commonly adopted in video pose estimation. Following prior work \citep{zeng2022smoothnet,jin2023kinematic}, we compute an acceleration-based jitter score ($J_{\text{acc}}$) as the mean magnitude of the second-order temporal difference of 2D keypoints, averaged over joints and time. To capture sharper temporal instabilities, we additionally compute a jerk-based score ($J_{\text{jerk}}$) using the third-order temporal difference, which penalizes rapid changes in acceleration and has been used as a smoothness indicator in motion and inertial pose estimation \citep{yi2021transpose,di2022sensors}. We also report motion energy ($E_v$), defined as the mean joint velocity magnitude, to contextualize jitter values with respect to the overall amount of motion in the sequence. For 3D estimators, we compute these metrics using only the 2D image-plane coordinates.

Given the importance of manual and non-manual (e.g., facial) articulation in sign language, all metrics are computed on three subsets: (i) all keypoints (excluding legs), (ii) hands only, and (iii) face only. For each sequence, derivative magnitudes are averaged over joints and over time, yielding one scalar per sequence.
Evaluation is conducted on 20 randomly selected videos from Signsuisse and 20 from Phoenix. Results are reported as median and interquartile range (IQR) across sequences (instead of mean and standard deviation), as these are more robust to the right-skewed nature of the observed distributions. Lower $J_{\text{acc}}$ and $J_{\text{jerk}}$ indicate smoother and more temporally stable predictions.

\paragraph{Occlusion} Occlusion is a common source of error in pose estimation \citep{lino2025benchmarking,fan2025pose}, particularly in sign language, where one hand frequently overlaps the other or the face. To assess how estimators behave under these conditions, we analyzed 10 hand-picked videos from the Signsuisse dataset, comprising 15 individual signs that include occlusion.

We define \textit{occlusion} as frames in which one hand or arm partially or fully obstructs the other hand or arm from the camera’s viewpoint. We visually inspected the pose outputs for each estimator and evaluated whether both hands were consistently detected and whether their locations, handshapes, movements, and palm orientations remained plausible relative to the source video. A prediction was considered acceptable when both hands were present for most of the sequence and their articulatory configuration matched the observed signing. Given the limited sample size, this occlusion analysis is intended to contextualize the main translation results rather than provide a comprehensive robustness benchmark. 

\paragraph{Missing hand keypoints}
During translation preprocessing, keypoints with confidence $c=0$ are treated as invalid and zero-filled before frame-level flattening; such keypoints are therefore absent from the model input. Estimators that frequently produce $c=0$ hand keypoints thus provide less hand information to the translation model. Example cases are shown in Figure~\ref{fig:pose_comparison}. We therefore quantify missing hand keypoints directly from the \texttt{.pose} files to capture how often substantial portions of hand information are unavailable to the translation model.

For each frame and each hand, we compute the proportion of hand keypoints with $c=0$. We define a hand as \textit{missing} in a given frame when at least $50\%$ of its keypoints have $c=0$. Signing frames are defined as those where the wrist is vertically above the elbow.

\begin{table}
    \centering
    \resizebox{\linewidth}{!}{
    \begin{tabular}{lrr}
         \toprule
         \textbf{Estimator} & \textbf{BLEU ($\uparrow$)} & \textbf{BLEURT ($\uparrow$)} \\
         \midrule
        MediaPipe & $10.327$ $\pm$ $0.269$ & $0.351$ $\pm$ $0.005$ \\
        OpenPose & $10.606$ $\pm$ $0.251$ & $0.353$ $\pm$ $0.002$ \\
        MMPose Wholebody & $10.901$ $\pm$ $0.299$ & $0.361$ $\pm$ $0.007$ \\
        OpenPifPaf & $9.365$ $\pm$ $0.263$ & $0.325$ $\pm$ $0.007$ \\
        SDPose & $11.681$ $\pm$ $0.415$ & $0.372$ $\pm$ $0.001$ \\
        Sapiens & $11.525$ $\pm$ $0.222$ & $0.372$ $\pm$ $0.003$ \\
        AlphaPose & $11.251$ $\pm$ $0.241$ & $0.359$ $\pm$ $0.007$ \\
        SMPLest-X & $9.709$ $\pm$ $0.334$ & $0.341$ $\pm$ $0.009$ \\
         \bottomrule
    \end{tabular}
    }
    \caption{Translation scores on the Phoenix dataset. We report the mean and standard deviation across three training runs.}
    \label{tab:translation_results}
\end{table}

\section{Results \& Discussion}

\subsection{Translation Results}

Table \ref{tab:translation_results} shows the evaluation results of our sign language translation (SLT) experiments. Every result is an average across three training runs. Overall, the translation quality in our experiments is much lower than state-of-the art systems, especially ones based on videos instead of poses, but we are interested specifically in the \textit{relative} performance of pose estimators compared to each other.

Sapiens and SDPose achieved the highest BLEU and BLEURT scores, of roughly $11.5$ and $0.37$, suggesting that they are stronger candidates for SLT applications compared to other estimators. In total, four different estimators (Sapiens, SDPose, MMPose Wholebody and AlphaPose) lead to better results than Mediapipe, the de-facto standard pose estimation system used in SLT.

This provides empirical evidence that \textbf{alternatives to Mediapipe may be better suited for SLT}, and \textbf{future work should therefore consider a broader range of pose estimation systems.}


\subsection{Further Analyses}

\begin{figure}
    \centering
    \setlength{\abovecaptionskip}{2pt}
    \setlength{\belowcaptionskip}{-6pt}
    
    \includegraphics[trim={90pt 167pt 163pt 241pt}, clip, width=0.49\linewidth]
    {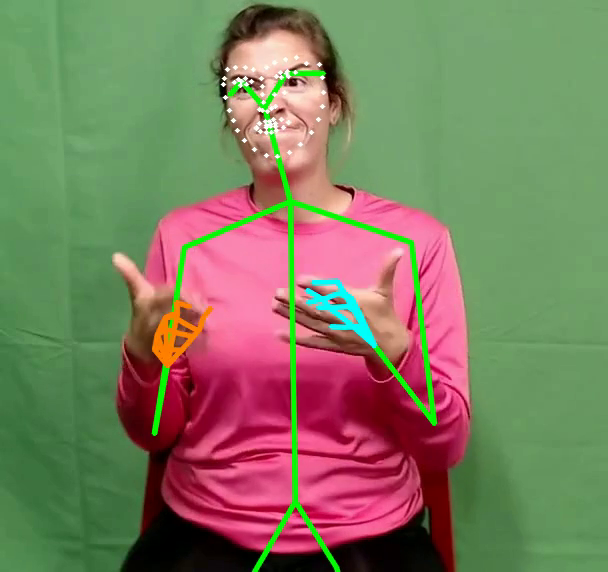}
    \hfill
    \includegraphics[trim={140pt 107pt 133pt 201pt}, clip, width=0.49\linewidth]
    {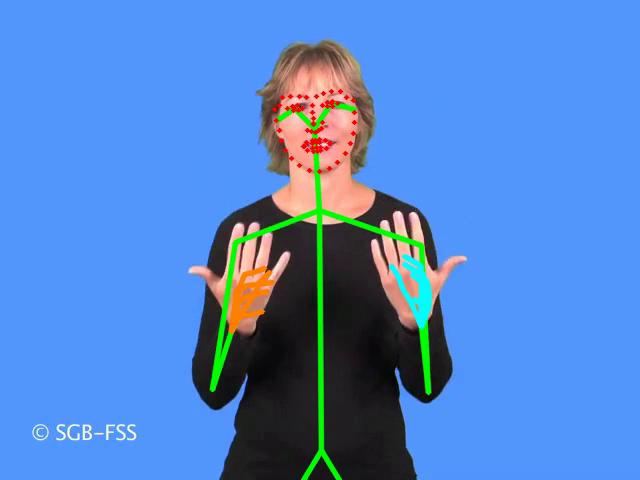}
    
    \caption{Examples of erroneous SMPLest-X hand pose estimates overlaid on original frames, cropped to highlight the hands. In both examples (left: How2Sign \citep{Duarte_2021_CVPR}; right: Signsuisse), the predicted hand keypoints exhibit incorrect orientation and articulation that do not match the observed hand configurations.}

    \label{fig:smplest_x_how2sign_signswiss_side}
\end{figure}

\begin{table*}[t]
\centering
\small
\setlength{\tabcolsep}{4pt}

\begin{tabular*}{\textwidth}{l@{\extracolsep{\fill}}ccc ccc}
\toprule
& \multicolumn{3}{c}{\textbf{Signsuisse}} 
& \multicolumn{3}{c}{\textbf{Phoenix}} \\
\cmidrule(lr){2-4} \cmidrule(lr){5-7}
\textbf{Pose Estimator}
& $E_v$ & $J_{\text{acc}}$ ($\downarrow$) & $J_{\text{jerk}}$ ($\downarrow$)
& $E_v$ & $J_{\text{acc}}$ ($\downarrow$) & $J_{\text{jerk}}$ ($\downarrow$) \\
\midrule

MediaPipe
& $\mathbf{1.16}$ & $\mathbf{1.44}$ \tiny{(1.13--1.60)} & $\mathbf{2.46}$ \tiny{(2.01--2.81)}
& $\mathbf{3.07}$ & $3.69$ \tiny{(3.01--4.73)} & $6.51$ \tiny{(5.10--8.38)} \\

OpenPose
& $3.30$ & $3.97$ \tiny{(3.56--4.74)} & $6.84$ \tiny{(5.90--8.18)}
& $9.32$ & $14.74$ \tiny{(11.38--16.64)} & $26.36$ \tiny{(19.57--29.55)} \\

MMPose Wholebody
& $3.83$ & $4.95$ \tiny{(4.26--5.90)} & $8.58$ \tiny{(7.32--10.35)}
& $6.84$ & $8.75$ \tiny{(7.00--11.17)} & $14.47$ \tiny{(11.66--18.59)} \\

OpenPifPaf
& $2.76$ & $4.84$ \tiny{(3.43--5.54)} & $8.94$ \tiny{(6.35--10.20)}
& $5.22$ & $8.60$ \tiny{(7.28--10.31)} & $15.35$ \tiny{(13.12--18.63)} \\

SDPose
& $6.21$ & $8.45$ \tiny{(6.19--11.76)} & $14.97$ \tiny{(10.70--21.79)}
& $6.83$ & $7.99$ \tiny{(6.61--10.45)} & $13.24$ \tiny{(10.80--17.10)} \\

Sapiens
& $2.97$ & $4.42$ \tiny{(3.72--5.05)} & $7.77$ \tiny{(6.62--9.16)}
& $4.55$ & $6.33$ \tiny{(5.44--7.43)} & $10.86$ \tiny{(9.62--12.88)} \\

AlphaPose
& $4.09$ & $4.56$ \tiny{(3.08--6.70)} & $7.36$ \tiny{(5.15--11.50)}
& $5.81$ & $6.21$ \tiny{(4.90--7.69)} & $10.32$ \tiny{(8.35--12.47)} \\

SMPLest-X
& $2.13$ & $1.74$ \tiny{(1.39--2.00)} & $2.74$ \tiny{(2.16--3.14)}
& $3.08$ & $\mathbf{2.68}$ \tiny{(2.23--3.30)} & $\mathbf{3.97}$ \tiny{(3.19--4.85)} \\

\bottomrule
\end{tabular*}

\caption{
Temporal stability metrics on Signsuisse and Phoenix datasets, for all keypoints excluding legs.
$E_v$: motion energy (median joint velocity).
$J_{\text{acc}}$ and $J_{\text{jerk}}$: acceleration- and jerk-based jitter; lower is smoother.
Values are median across 20 sequences, with interquartile range (Q1--Q3) in parentheses.
All values scaled by $100$.
Boldface marks the lowest value when distributions are clearly separated (see Figure~\ref{fig:jitter_violin_all}).
Full results by anatomical region in Appendix~\ref{app:jitter_full}.
}
\label{tab:jitter_main}
\end{table*}

\begin{figure*}[t]
    \centering
    \begin{subfigure}[t]{0.49\textwidth}
        \centering
        \includegraphics[width=\linewidth]{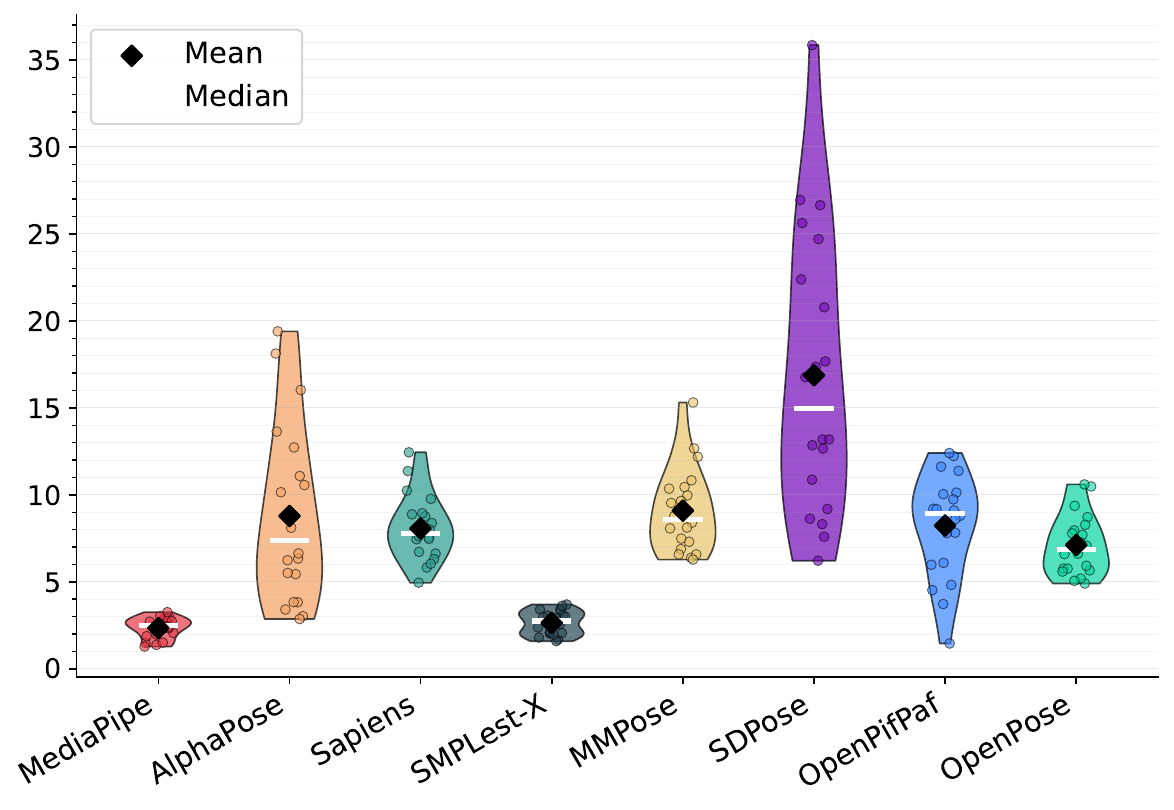}
        \caption{Signsuisse}
        \label{fig:jitter_violin_signsuisse}
    \end{subfigure}
    \hfill
    \begin{subfigure}[t]{0.49\textwidth}
        \centering
        \includegraphics[width=\linewidth]{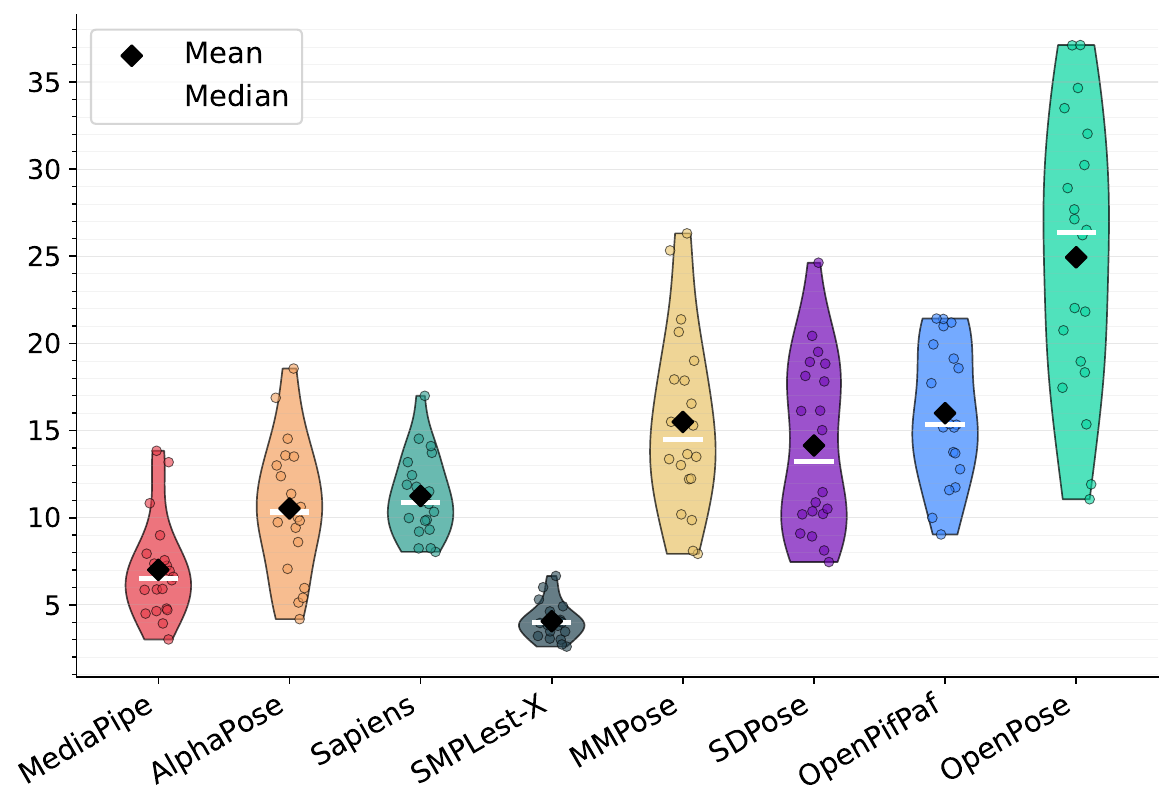}
        \caption{Phoenix}
        \label{fig:jitter_violin_phoenix}
    \end{subfigure}
 
    \caption{Distribution of per-sequence jerk jitter ($J_{\text{jerk}}$) across pose estimators for all keypoints (excluding legs). Each point represents one video sequence ($n=20$). Black diamonds indicate the mean; white bars indicate the median. The violin shape shows the kernel density estimate. Overlapping distributions (e.g., MediaPipe vs.\ SMPLest-X on Signsuisse) indicate that differences between estimators may not be reliable at this sample size.}
    \label{fig:jitter_violin_all}
\end{figure*}

\paragraph{Temporal jittering (pose stability)}

Table~\ref{tab:jitter_main} summarizes the temporal stability of pose estimators on Signsuisse and Phoenix for all keypoints (excluding legs); full results by anatomical region (hands, face) are in Appendix~\ref{app:jitter_full}. Overall, the quantitative trends align with visual inspection: methods that appear temporally unstable exhibit higher $J_{\text{acc}}$ and $J_{\text{jerk}}$, particularly in the hand subset. While both metrics exhibit similar trends, $J_{\text{jerk}}$ provides greater separation between estimators (Figure~\ref{fig:jitter_violin_all}) and is thus used to illustrate differences.
 
On Signsuisse, MediaPipe achieves the lowest median jerk jitter for all keypoints ($J_{\text{jerk}}=2.46$), followed closely by SMPLest-X ($2.74$). However, the violin plots reveal that their distributions overlap substantially (Figure~\ref{fig:jitter_violin_signsuisse}), indicating that the difference may not be reliable given the sample size of 20 sequences. SDPose exhibits the highest jerk values ($14.97$), with a wide interquartile range reflecting high variance across sequences, despite yielding the most stable face keypoints (see Appendix Table~\ref{tab:jitter_appendix_face} for full results).

When focusing on hands (Appendix Table~\ref{tab:jitter_appendix_hands}), SMPLest-X provides the lowest jitter ($J_{\text{jerk}}=5.85$); however, this temporal smoothness coincides with visibly rigid hand configurations (Figure~\ref{fig:smplest_x_how2sign_signswiss_side}). For face keypoints, SDPose achieves the lowest jitter ($0.38$), though MediaPipe ($0.44$) and OpenPifPaf ($0.43$) are nearly indistinguishable.
 
On Phoenix, instability increases for all methods, consistent with the lower spatial resolution of the videos (approximately $5.3\times$ fewer pixels than Signsuisse). SMPLest-X achieves the lowest median jerk jitter for all keypoints ($3.97$) and hands ($7.40$), with distributions clearly separated from other estimators (Figure~\ref{fig:jitter_violin_phoenix}). MediaPipe is second-lowest overall ($6.51$). OpenPose exhibits the highest jitter across all regions, with a median $J_{\text{jerk}}$ of $26.36$ for all keypoints and $67.74$ for hands. For face keypoints, SDPose achieves the lowest jitter ($1.17$), while Sapiens is notably high ($10.29$), consistent with its dense 243-landmark face mesh amplifying localization noise at low resolution.
 
Across both datasets, the violin plots (Figure~\ref{fig:jitter_violin_all}) provide important context beyond summary statistics: estimators with similar median values may have heavily overlapping distributions, tempering conclusions about which is ``best.'' When interpreted jointly with motion energy, the results suggest that SMPLest-X favors temporal smoothness at the cost of limited articulation, while MediaPipe---and to a lesser extent Sapiens---offer a trade-off between stability and expressive motion.

\begin{figure}
    \centering
    \setlength{\abovecaptionskip}{2pt}
    \setlength{\belowcaptionskip}{-6pt}
    
    \includegraphics[trim={155pt 0pt 150pt 0pt}, clip, width=0.47\linewidth]
    {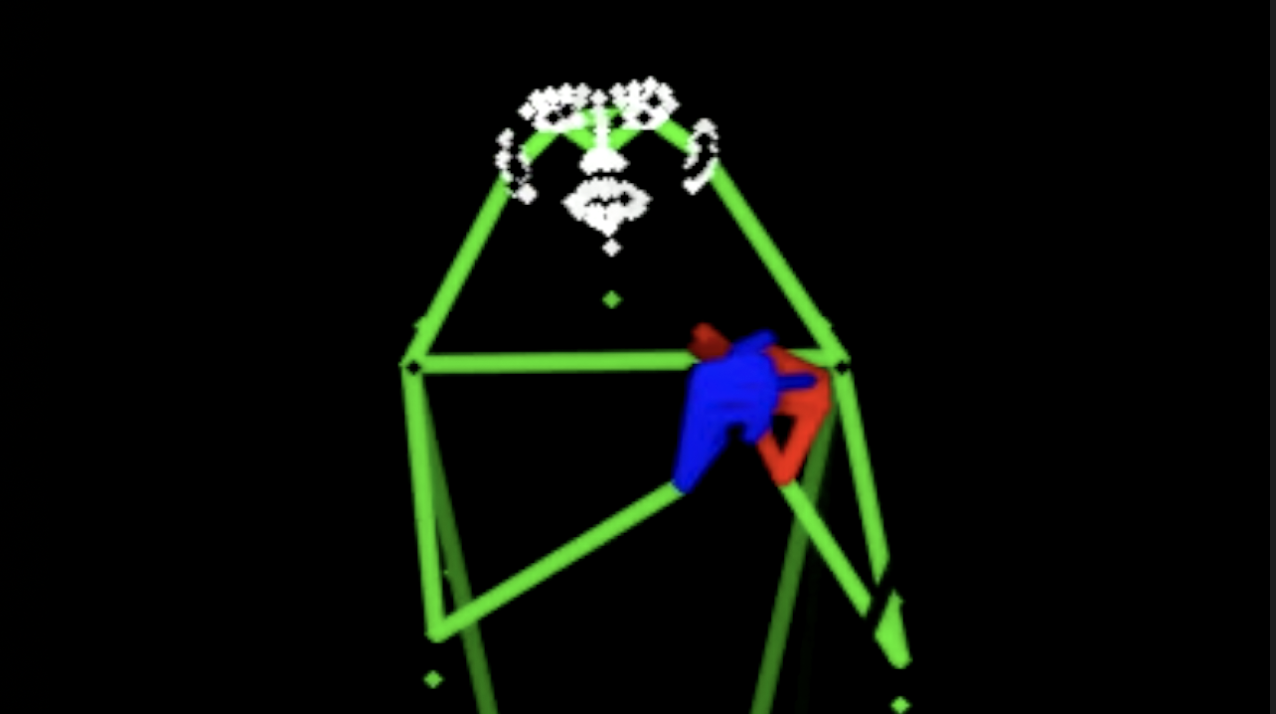}
    \hfill
    \includegraphics[trim={155pt 0pt 150pt 0pt}, clip, width=0.463\linewidth]
    {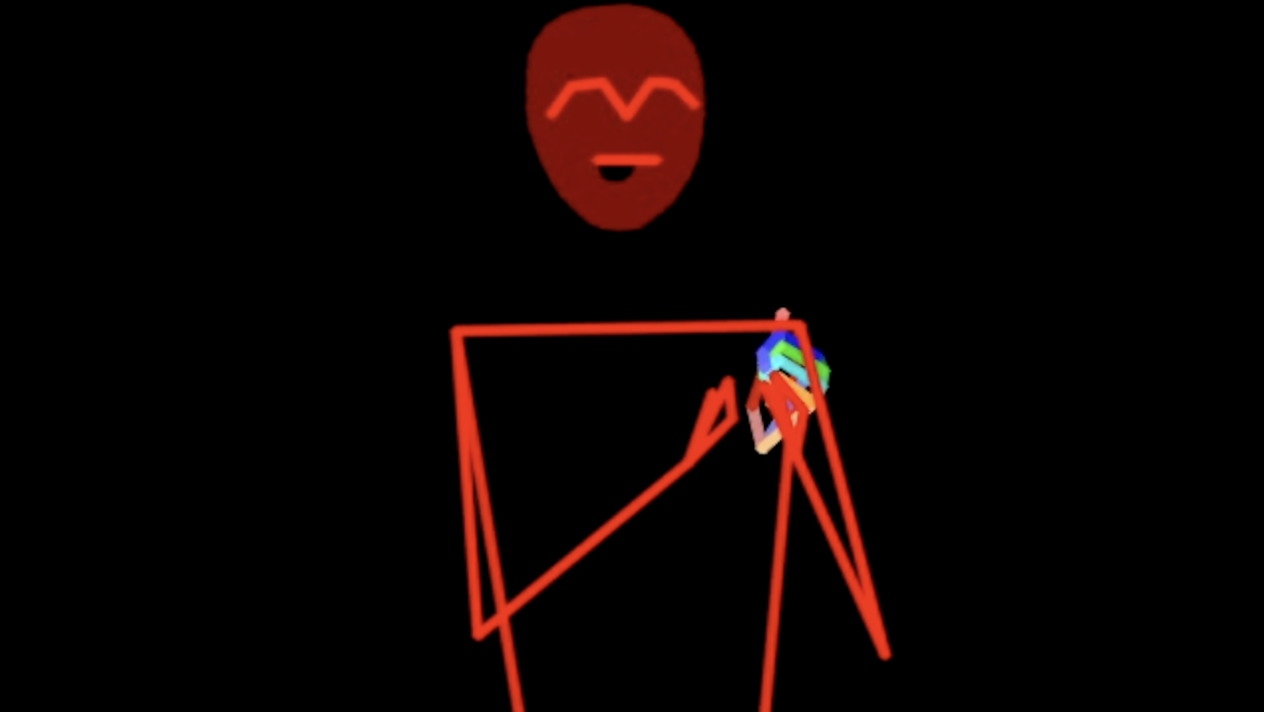}
    
    \caption{Pose estimation visualizations with hand occlusion for the DSGS sign \texttt{SPORT}. Left: Sapiens correctly predicts both hands despite partial occlusion. Right: MediaPipe fails to detect the right hand, resulting in missing keypoints. Images are cropped to highlight the hands.}
    \label{fig:occlusion-figure}
\end{figure}

\paragraph{Occlusion}
The results of the occlusion analysis are available in Table~\ref{tab:Occlusion-Table}.  Sapiens produced accurate estimations for 15/15 instances of occlusion surveyed. Notably, there are several circumstances where MediaPipe is missing one or both hands for a given sign, but Sapiens produces the pose correctly. One example of this trend is shown in Figure ~\ref{fig:occlusion-figure}. 
OpenPifPaf, on the other hand, was missing some or all of one or both hands in all but one of the 15 instances of occlusion surveyed. This propensity for missing hands may contribute to the low BLEU score of OpenPifPaf. SMPLest-X, in turn, did provide hands in all frames, but did not produce any accurate representations of the pose. 

MMPose Wholebody and SDPose both estimated a hand position for all frames in all of the 15 examples surveyed, but were more likely than other estimators to estimate an impossible hand position, for example bending the fingers backwards. MMPoseWholebody created an acceptable estimation for 5/15 instances of occlusion, while SDPose produced an acceptable estimation in 7/15 examples.

\begin{table}
    \centering
    \begin{tabular*}{\linewidth}{l@{\extracolsep{\fill}}r}\toprule
         \textbf{estimator}& \textbf{correctness (\%)}\\\midrule
         Mediapipe & $73.33$ \\
         OpenPose & $66.66$ \\
         MMPose Wholebody & $33.33$ \\
         OpenPifPaf & $6.66$ \\
         SDPose & $46.66$ \\
         Sapiens & $100.00$ \\
         AlphaPose-136 & $40.00$ \\
         SMPLest-X & $0.00$ \\ \bottomrule
    \end{tabular*}
    \caption{Percentage of the time that the estimators correctly estimated a pose featuring occlusion in the 15 Signsuisse examples surveyed.}
    \label{tab:Occlusion-Table}
\end{table}

\begin{table}[t]
\centering
\begin{tabular*}{\linewidth}{l@{\extracolsep{\fill}}rrr}
\toprule
\textbf{estimator} & \textbf{left} & \textbf{right} & \textbf{both} \\
\midrule
MediaPipe  & $20.22$ & $23.61$ & $8.84$  \\
OpenPose   & $8.43$  & $3.88$  & $0.22$  \\
OpenPifPaf & $67.65$ & $59.19$ & $40.63$ \\
\bottomrule
\end{tabular*}
\caption{Percentage of signing frames with a missing hand on 20 randomly selected Phoenix training videos. A hand is missing when $\geq 50\%$ of its keypoints have $c=0$. ``both'': both hands missing simultaneously. Only estimators producing $c=0$ hand keypoints are shown; all others (AlphaPose, MMPose Wholebody, Sapiens, SDPose, SMPLest-X) score $0\%$. Full threshold sweep in Appendix~\ref{app:missing_hands}.}
\label{tab:missing_hands}
\end{table}

\paragraph{Missing hand keypoints}
We report the percentage of signing frames with a missing hand for each estimator (Table~\ref{tab:missing_hands}), computed on 20 randomly selected videos from the Phoenix training split. A full breakdown across varying missing thresholds ($10$--$100\%$) is provided in Appendix~\ref{app:missing_hands}. Only OpenPifPaf, MediaPipe, and OpenPose assign $c=0$ to hand keypoints; all other estimators never produce $c=0$, resulting in $0\%$ missing across all thresholds.
 
OpenPifPaf shows severe degradation: in roughly $68\%$ of signing frames the left hand is missing and in $59\%$ the right hand is missing. In over $40\%$ of signing frames, both hands are simultaneously absent. MediaPipe exhibits a milder but consistent pattern, with around $20$--$24\%$ of signing frames missing a hand. OpenPose is affected less frequently, with $3$--$8\%$ of frames affected. Notably, both MediaPipe and OpenPose exhibit binary behavior: when a hand is lost, all of its keypoints are set to $c=0$ at once rather than partially, as shown by the constant values across thresholds in the full breakdown (Table~\ref{tab:missing_hands_full} in Appendix \ref{app:missing_hands}).
 
These three estimators also obtain some of the weakest translation scores (Table~\ref{tab:translation_results}), suggesting that zero-confidence hand keypoints---explicitly masked during preprocessing---directly impair downstream translation. This pattern could also reveal a limited capability of current translation models to recover missing hand information from temporal context alone.


\section{Conclusions}


We presented a controlled comparison of pose estimators for pose-based SLT, motivated by the fact that most prior SLT pipelines default to MediaPipe as a convenient choice. Our experiments show that this default is not necessarily optimal: several estimators outperform MediaPipe on Phoenix, including SDPose, Sapiens, AlphaPose, and MMPose Wholebody. Among them, SDPose and Sapiens achieve the highest BLEU/BLEURT scores.

\textbf{Further analyses on signing data}
Estimators that frequently remove substantial hand information (notably OpenPifPaf, and to a lesser extent MediaPipe and OpenPose) are associated with weaker translation results. The temporal stability analysis further shows that higher translation performance does not necessarily coincide with smoother trajectories: MediaPipe and SMPLest-X yield the lowest jitter scores, whereas SDPose exhibits comparatively high hand jerk on Signsuisse, and SMPLest-X attains very low jitter at the cost of visibly implausible or rigid hand articulations. Taken together, these findings underscore that estimator choice can materially affect both translation performance and pose quality characteristics relevant to signing.

\textbf{Trade-off between translation quality and computational cost} Sapiens provides strong translation performance and the most robust behavior under occlusion (15/15 correct in our manual analysis), but it is also among the most resource-intensive and slowest options. SDPose reaches comparable translation quality results with similarly high runtime requirements. Conversely, faster or more lightweight estimators are not guaranteed to be suitable for SLT: OpenPifPaf, for instance, performs poorly in translation and exhibits severe hand keypoint dropout. AlphaPose, which has the third-best BLEU score and the fastest compute speed, may be a better option when time or computing resources are limited. 

Overall, our results provide empirical evidence that pose-based SLT should not treat pose estimation as a fixed implementation detail. Future research should therefore consider pose estimators beyond MediaPipe when building pose-based translation systems, and evaluate estimator choice jointly with practical constraints such as runtime and hardware requirements as well as robustness properties such as occlusion handling, missing-hand behavior, and temporal stability.

\section{Limitations and Future Work}



\textbf{Limitations of the Phoenix dataset} It should be noted that the signing in the Phoenix dataset is done live by hearing interpreters. Accordingly, this dataset has notable flaws. Due to the time pressure of the live setting, the interpreters may omit some information. Furthermore, the signing is an interpretation of German spoken language, and not natural signing, and thus may be influenced by German grammar. Lastly, the subject domain of Phoenix is fairly limited and mostly pertains to weather reports. Future investigations should expand the analysis of pose estimators to new datasets, particularly those that contain natural signing from deaf L1 signers with high-quality translations that are not produced under time pressure.

Translation experiments on more suitable, newer, or larger sign language datasets would provide a more reliable assessment of how pose estimator choice generalizes beyond Phoenix.

\textbf{Fast movement}
Anecdotally, a visual analysis suggests that estimators such as Sapiens may be vulnerable to errors on videos containing fast movements (which is normal for signing). Future research could therefore investigate the estimators' robustness to fast movements.

\textbf{Variable model capacity}
All experiments use a shared translation architecture with a linear projection layer that maps pose features into the language model embedding space (see Section~\ref{subsec:translation_method}). Because pose estimators produce different numbers of keypoints (Table~\ref{tab:estimators_comparison}), this projection layer contains a different number of parameters for each estimator. While this design allows each representation to be fully utilized, it introduces variation in model capacity that may influence translation performance. Isolating the impact of adapter dimensionality is left for future work.



\section*{Acknowledgements}

CO is supported by a Fulbright Scholarship.
GS and MM received funding from the SIGMA project (grant no. G-95017-01-07), supported by the Digital Society Initiative (DSI) at the University of Zurich.




\section{Bibliographical References}\label{sec:reference}

\bibliographystyle{lrec2026-natbib}
\bibliography{lrec2026-example}

\clearpage
\appendix

\twocolumn[%
  \vspace{0.5em}
  \centering
  \footnotesize
  \setlength{\tabcolsep}{3.5pt}
  \begin{tabular*}{\textwidth}{ll@{\extracolsep{\fill}} rrrrrrrrrr}
  \toprule
  & & \multicolumn{10}{c}{\textbf{\% hand keypoints missing ($\geq x$)}} \\
  \cmidrule(lr){3-12}
  \textbf{Estimator} & \textbf{Hand} 
    & \textbf{10\%} & \textbf{20\%} & \textbf{30\%} & \textbf{40\%} & \textbf{50\%} 
    & \textbf{60\%} & \textbf{70\%} & \textbf{80\%} & \textbf{90\%} & \textbf{100\%} \\
  \midrule
  \multirow{3}{*}{OpenPifPaf}
    & Left  & $85.16$ & $80.86$ & $76.51$ & $72.10$ & $67.65$ & $63.29$ & $59.31$ & $54.59$ & $51.81$ & $0.00$ \\
    & Right & $82.59$ & $75.01$ & $69.90$ & $63.68$ & $59.19$ & $55.66$ & $51.96$ & $49.01$ & $46.19$ & $0.00$ \\
    & Both  & $66.90$ & $57.98$ & $52.14$ & $45.91$ & $40.63$ & $36.43$ & $32.32$ & $28.91$ & $26.05$ & $0.00$ \\
  \midrule
  \multirow{3}{*}{MediaPipe}
    & Left  & $20.22$ & $20.22$ & $20.22$ & $20.22$ & $20.22$ & $20.22$ & $20.22$ & $20.22$ & $20.22$ & $0.00$ \\
    & Right & $23.61$ & $23.61$ & $23.61$ & $23.61$ & $23.61$ & $23.61$ & $23.61$ & $23.61$ & $23.61$ & $0.00$ \\
    & Both  & $8.84$  & $8.84$  & $8.84$  & $8.84$  & $8.84$  & $8.84$  & $8.84$  & $8.84$  & $8.84$  & $0.00$ \\
  \midrule
  \multirow{3}{*}{OpenPose}
    & Left  & $8.43$  & $8.43$  & $8.43$  & $8.43$  & $8.43$  & $8.43$  & $8.43$  & $8.43$  & $8.43$  & $0.00$ \\
    & Right & $3.88$  & $3.88$  & $3.88$  & $3.88$  & $3.88$  & $3.88$  & $3.88$  & $3.88$  & $3.88$  & $0.00$ \\
    & Both  & $0.22$  & $0.22$  & $0.22$  & $0.22$  & $0.22$  & $0.22$  & $0.22$  & $0.22$  & $0.22$  & $0.00$ \\
  \bottomrule
  \end{tabular*}
  \captionof{table}{Percentage of signing frames in which at least $x\%$ of a hand's keypoints have confidence $c=0$, measured on 20 randomly selected videos from the Phoenix training split. Signing frames are those where the wrist is vertically above the elbow. ``Both'': both hands simultaneously exceed the threshold. Only estimators producing $c=0$ hand keypoints are shown; all others score $0\%$. The $50\%$ column corresponds to the missing hand definition in Table~\ref{tab:missing_hands}.}
  \label{tab:missing_hands_full}
  \vspace{1em}
]
\section{Missing Hand Keypoints}\label{app:missing_hands}
Table~\ref{tab:missing_hands_full} reports the percentage of signing frames with missing hand keypoints across varying thresholds. The $50\%$ column corresponds to the definition of a missing hand used in the main text (Table~\ref{tab:missing_hands}). MediaPipe and OpenPose exhibit binary behavior: when a hand is lost, all its keypoints are set to $c=0$ simultaneously, resulting in constant values across thresholds $10$--$90\%$.

\begin{table*}[t]
\centering
\small
\setlength{\tabcolsep}{4pt}
\begin{tabular*}{\textwidth}{l@{\extracolsep{\fill}}ccc ccc}
\toprule
& \multicolumn{3}{c}{\textbf{Signsuisse}}
& \multicolumn{3}{c}{\textbf{Phoenix}} \\
\cmidrule(lr){2-4} \cmidrule(lr){5-7}
\textbf{Pose Estimator}
& $E_v$ & $J_{\text{acc}}$ ($\downarrow$) & $J_{\text{jerk}}$ ($\downarrow$)
& $E_v$ & $J_{\text{acc}}$ ($\downarrow$) & $J_{\text{jerk}}$ ($\downarrow$) \\
\midrule
MediaPipe
& $10.49$ & $14.50$ \tiny{(10.73--16.23)} & $25.29$ \tiny{(20.01--29.52)}
& $15.99$ & $25.12$ \tiny{(20.58--31.66)} & $45.99$ \tiny{(35.73--56.71)} \\
OpenPose
& $9.78$ & $11.64$ \tiny{(10.01--13.55)} & $20.18$ \tiny{(17.39--23.69)}
& $23.47$ & $37.63$ \tiny{(27.70--45.74)} & $67.74$ \tiny{(47.80--82.59)} \\
MMPose Wholebody
& $7.01$ & $6.54$ \tiny{(4.78--8.47)} & $10.48$ \tiny{(7.81--14.08)}
& $12.28$ & $12.26$ \tiny{(9.10--17.67)} & $19.58$ \tiny{(14.08--28.58)} \\
OpenPifPaf
& $6.88$ & $11.96$ \tiny{(8.00--14.38)} & $22.61$ \tiny{(15.11--26.69)}
& $13.66$ & $24.46$ \tiny{(19.96--28.12)} & $44.35$ \tiny{(36.52--51.38)} \\
SDPose
& $13.32$ & $16.53$ \tiny{(11.68--25.10)} & $29.98$ \tiny{(20.35--45.73)}
& $12.31$ & $12.62$ \tiny{(8.47--17.52)} & $18.37$ \tiny{(13.62--26.73)} \\
Sapiens
& $8.27$ & $8.42$ \tiny{(6.39--14.92)} & $13.61$ \tiny{(10.67--27.18)}
& $10.01$ & $9.94$ \tiny{(7.44--15.42)} & $15.27$ \tiny{(11.92--26.14)} \\
AlphaPose
& $12.24$ & $13.53$ \tiny{(7.72--21.19)} & $21.81$ \tiny{(12.82--36.93)}
& $13.94$ & $16.36$ \tiny{(11.23--19.74)} & $25.85$ \tiny{(18.36--32.92)} \\
SMPLest-X
& $\mathbf{5.53}$ & $\mathbf{3.87}$ \tiny{(3.18--4.51)} & $\mathbf{5.85}$ \tiny{(4.72--6.80)}
& $\mathbf{7.36}$ & $\mathbf{5.56}$ \tiny{(4.68--7.02)} & $\mathbf{7.40}$ \tiny{(6.49--9.34)} \\
\bottomrule
\end{tabular*}
\caption{
Temporal stability metrics for hand keypoints on Signsuisse and Phoenix.
Format as in Table~\ref{tab:jitter_main}: median (IQR), scaled by $100$.
}
\label{tab:jitter_appendix_hands}
\end{table*}

\begin{table*}[t]
\centering
\small
\setlength{\tabcolsep}{4pt}
\begin{tabular*}{\textwidth}{l@{\extracolsep{\fill}}ccc ccc}
\toprule
& \multicolumn{3}{c}{\textbf{Signsuisse}}
& \multicolumn{3}{c}{\textbf{Phoenix}} \\
\cmidrule(lr){2-4} \cmidrule(lr){5-7}
\textbf{Pose Estimator}
& $E_v$ & $J_{\text{acc}}$ ($\downarrow$) & $J_{\text{jerk}}$ ($\downarrow$)
& $E_v$ & $J_{\text{acc}}$ ($\downarrow$) & $J_{\text{jerk}}$ ($\downarrow$) \\
\midrule
MediaPipe
& $0.31$ & $0.27$ \tiny{(0.24--0.31)} & $0.44$ \tiny{(0.38--0.49)}
& $1.83$ & $1.65$ \tiny{(1.18--2.85)} & $2.73$ \tiny{(2.07--4.81)} \\
OpenPose
& $0.39$ & $0.46$ \tiny{(0.42--0.50)} & $0.81$ \tiny{(0.73--0.87)}
& $3.07$ & $4.15$ \tiny{(2.47--5.73)} & $7.16$ \tiny{(4.26--10.32)} \\
MMPose Wholebody
& $0.44$ & $0.64$ \tiny{(0.58--0.67)} & $1.14$ \tiny{(1.04--1.19)}
& $1.63$ & $1.23$ \tiny{(1.08--1.50)} & $1.84$ \tiny{(1.63--2.20)} \\
OpenPifPaf
& $0.32$ & $0.27$ \tiny{(0.23--0.30)} & $0.43$ \tiny{(0.38--0.48)}
& $1.64$ & $1.08$ \tiny{(0.87--1.60)} & $1.52$ \tiny{(1.20--2.40)} \\
SDPose
& $0.31$ & $0.24$ \tiny{(0.22--0.27)} & $0.38$ \tiny{(0.34--0.42)}
& $1.59$ & $\mathbf{0.86}$ \tiny{(0.77--1.15)} & $\mathbf{1.17}$ \tiny{(1.01--1.42)} \\
Sapiens
& $2.20$ & $3.63$ \tiny{(3.17--4.08)} & $6.64$ \tiny{(5.76--7.45)}
& $3.92$ & $5.82$ \tiny{(5.23--6.56)} & $10.29$ \tiny{(9.37--11.60)} \\
AlphaPose
& $0.42$ & $0.55$ \tiny{(0.48--0.58)} & $0.95$ \tiny{(0.83--1.01)}
& $1.75$ & $1.41$ \tiny{(1.08--2.37)} & $2.22$ \tiny{(1.68--3.87)} \\
SMPLest-X
& $0.69$ & $0.81$ \tiny{(0.65--0.99)} & $1.38$ \tiny{(1.10--1.69)}
& $\mathbf{1.37}$ & $1.49$ \tiny{(1.27--1.77)} & $2.50$ \tiny{(2.09--2.97)} \\
\bottomrule
\end{tabular*}
\caption{
Temporal stability metrics for face keypoints on Signsuisse and Phoenix.
Format as in Table~\ref{tab:jitter_main}: median (IQR), scaled by $100$.
}
\label{tab:jitter_appendix_face}
\end{table*}

\section{Temporal Stability by Region}\label{app:jitter_full}
 
\begin{figure}[t]
    \centering
    \begin{tikzpicture}
        \node[anchor=south west,inner sep=0] (img1) at (0,0)
            {\includegraphics[width=\linewidth]{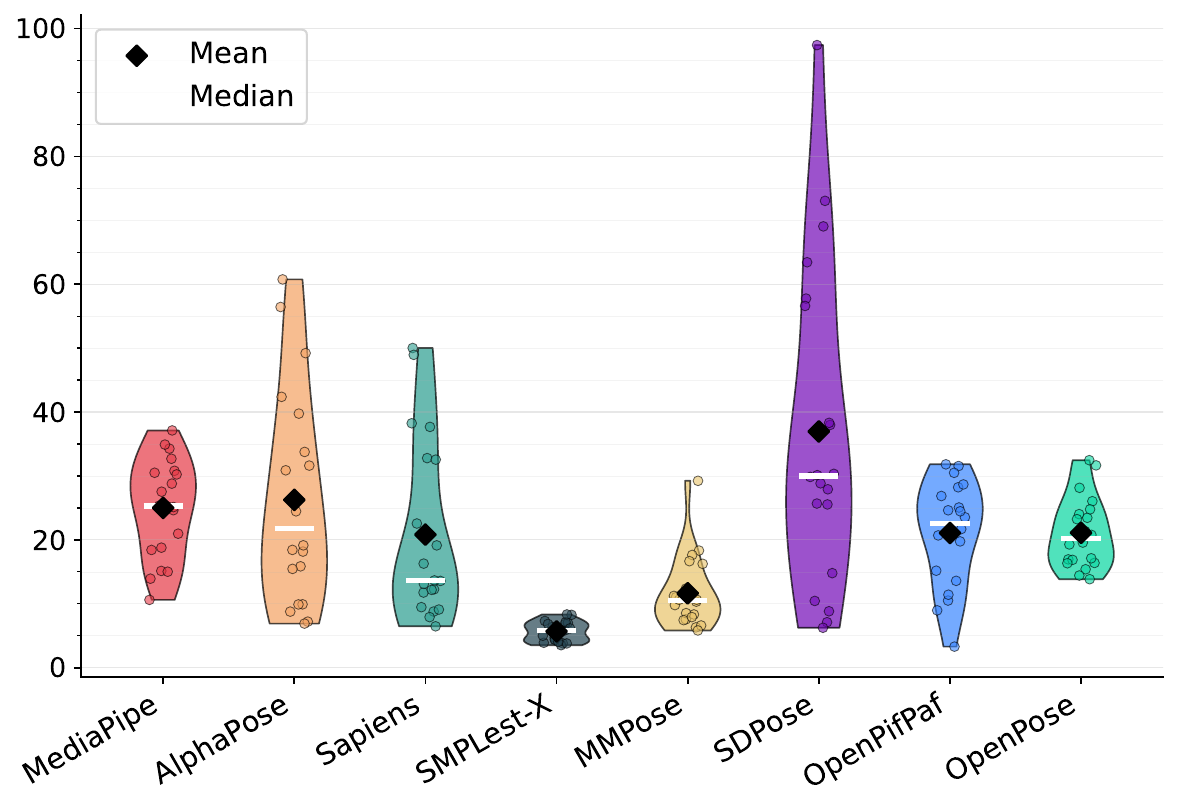}};
        \node[anchor=north,font=\normalfont\small,yshift=-4pt] at (img1.north) {Signsuisse};
    \end{tikzpicture}\par\vspace{-2pt}
    \begin{tikzpicture}
        \node[anchor=south west,inner sep=0] (img2) at (0,0)
            {\includegraphics[width=\linewidth]{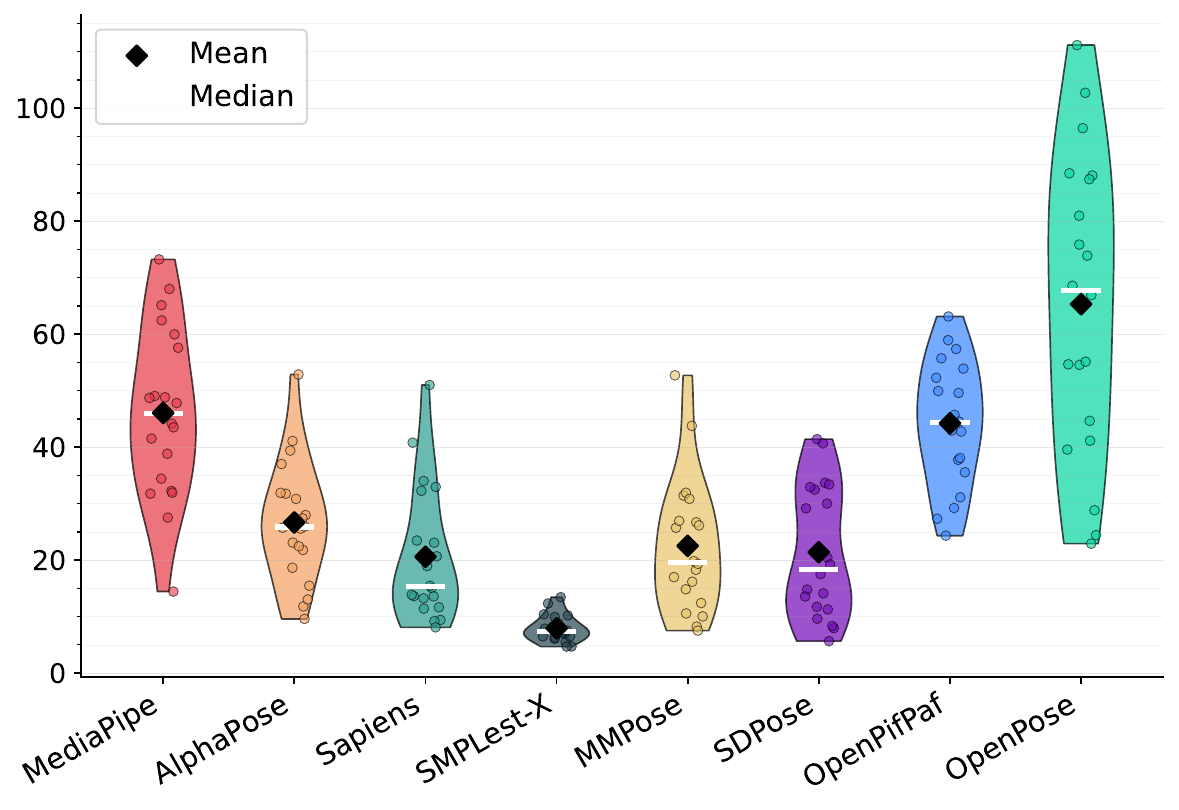}};
        \node[anchor=north,font=\normalfont\small,yshift=-4pt] at (img2.north) {Phoenix};
    \end{tikzpicture}\par\vspace{-4pt}
    \caption{Distribution of per-sequence jerk jitter ($J_{\text{jerk}}$) for hand keypoints on Signsuisse (top) and Phoenix (bottom). Format as in Figure~\ref{fig:jitter_violin_all}.}
    \label{fig:jitter_appendix_hands}
\end{figure}

\begin{figure}[H]
    \centering
    \begin{tikzpicture}
        \node[anchor=south west,inner sep=0] (img1) at (0,0)
            {\includegraphics[width=\linewidth]{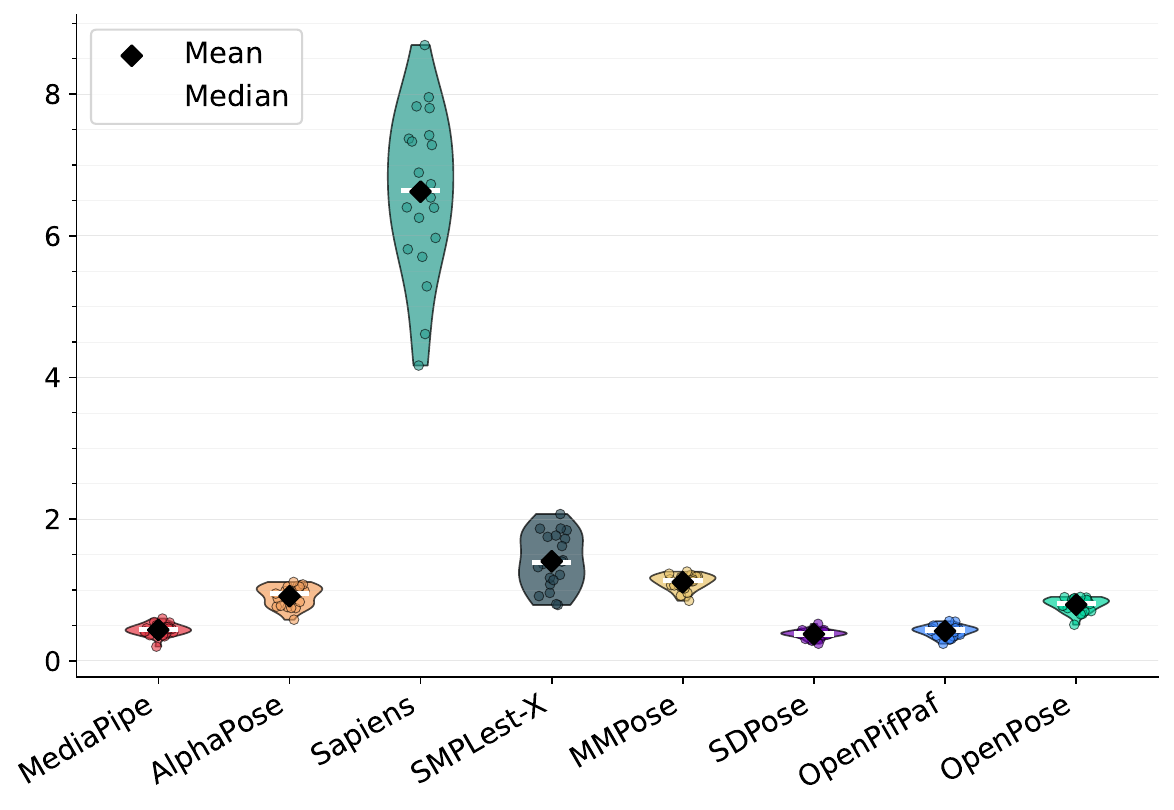}};
        \node[anchor=north,font=\normalfont\small,yshift=-4pt] at (img1.north) {Signsuisse};
    \end{tikzpicture}\par\vspace{-2pt}
    \begin{tikzpicture}
        \node[anchor=south west,inner sep=0] (img2) at (0,0)
            {\includegraphics[width=\linewidth]{phoenix_violin_jerk_mean_time_hands.pdf}};
        \node[anchor=north,font=\normalfont\small,yshift=-4pt] at (img2.north) {Phoenix};
    \end{tikzpicture}\par\vspace{-4pt}
    \caption{Distribution of per-sequence jerk jitter ($J_{\text{jerk}}$) for face keypoints on Signsuisse (top) and Phoenix (bottom). Format as in Figure~\ref{fig:jitter_violin_all}.}
    \label{fig:jitter_appendix_face}
\end{figure}

Tables~\ref{tab:jitter_appendix_hands} and~\ref{tab:jitter_appendix_face} report temporal stability metrics for hand and face keypoints on Signsuisse and Phoenix. The all-keypoints region is reported in the main text (Table~\ref{tab:jitter_main}). Corresponding violin plots are shown in Figures~\ref{fig:jitter_appendix_hands} and~\ref{fig:jitter_appendix_face}.

\paragraph{Hands}
Hand keypoints exhibit substantially higher jitter than full-body or face keypoints across all estimators, reflecting the difficulty of localizing fine-grained finger articulations.

On Signsuisse, SMPLest-X achieves the lowest median hand jerk jitter ($J_{\text{jerk}}=5.85$), clearly separated from the next-lowest estimator (MMPose Wholebody, $10.48$). However, as noted in the main text, this smoothness coincides with rigid, often implausible hand configurations (Figure~\ref{fig:smplest_x_how2sign_signswiss_side}). SDPose exhibits the highest hand jitter ($29.98$) with very wide interquartile range, indicating high variance across sequences. MediaPipe ($25.29$) and OpenPifPaf ($22.61$) also show high hand jitter on Signsuisse, though for different reasons: MediaPipe's hand jitter is inflated by its frequent missing-hand detections (Table~\ref{tab:missing_hands}), while OpenPifPaf's is driven by noisy localization even when hands are detected.
 
On Phoenix, the same pattern holds with larger values: SMPLest-X remains lowest ($7.40$), while OpenPose ($67.74$) and MediaPipe ($45.99$) are the most unstable. The increase relative to Signsuisse is consistent with Phoenix's lower resolution making hand keypoint localization more challenging.

\paragraph{Face}
Face keypoints show the lowest jitter overall, as facial landmarks undergo less motion than hands or body during signing. Differences between estimators are smaller but still informative.

On Signsuisse, SDPose ($J_{\text{jerk}}=0.38$), MediaPipe ($0.44$), and OpenPifPaf ($0.43$) are nearly indistinguishable and achieve the lowest face jitter. Sapiens is a clear outlier ($6.64$), likely because its dense 243-landmark face mesh amplifies small localization errors into measurable jitter.
 
On Phoenix, SDPose again achieves the lowest face jitter ($1.17$), with MMPose Wholebody ($1.84$) and OpenPifPaf ($1.52$) also performing well. Sapiens remains the highest ($10.29$), confirming that its dense face representation is especially sensitive to low-resolution input. MediaPipe shows moderate face jitter ($2.73$) with a wide IQR, suggesting inconsistent face tracking across sequences.


\end{document}